\pdfoutput=1
\documentclass[10pt,twocolumn,letterpaper]{article}
\usepackage[T1]{fontenc}
\usepackage{caption,algorithm,xcolor}
\usepackage[most]{tcolorbox}
\usepackage{cvpr}              
\usepackage{algpseudocode}

\usepackage{amsmath}
\usepackage{amssymb}

\graphicspath{{fig/}}

\usepackage{booktabs} 
\usepackage{multirow}
\newcommand{\G}{\mathcal{G}}
\newcommand{\V}{\mathcal{V}}
\newcommand{\E}{\mathcal{E}}
\newcommand{\A}{\mathbf{A}}
\definecolor{cvprblue}{rgb}{0.21,0.49,0.74}
\usepackage[breaklinks,colorlinks,allcolors=cvprblue]{hyperref}
\hypersetup{
  pdfinfo={
    Title={LLM-Guided Agentic Floor Plan Parsing for Accessible Indoor Navigation of Blind and Low-Vision People},
    Author={Aydin Ayanzadeh and Tim Oates}
  }
}

\def\confName{CVPR}
\def\confYear{2026}

\title{LLM-Guided Agentic Floor Plan Parsing for Accessible Indoor Navigation of Blind and Low-Vision People}

\author{Aydin Ayanzadeh and Tim Oates\\
University of Maryland, Baltimore County\\
{\tt\small \{aydina1, oates\}@umbc.edu}
}
\begin{document}
\maketitle
\begin{abstract}

Indoor navigation remains a critical accessibility challenge for blind and low-vision (BLV) individuals, as existing solutions rely on costly per-building infrastructure. We present an agentic framework that converts a single floor plan image into a structured, retrievable knowledge base to generate safe, accessible navigation instructions using lightweight infrastructure for BLV users. Our system has two main phases: a multi-agent system that constructs a knowledge graph by parsing the floorplan through a self-correcting pipeline, which employs iterative retry loops and corrective feedback to extract valid spatial knowledge graphs; and a Path Planner for generating accessible navigation instructions, evaluated by a Safety Evaluator to measure potential hazards during navigation. We evaluate our system in the real-world UMBC Math and Psychology building, floors 1 and 3. Our model surpasses baselines across all route types on MP-1 and MP-3. We achieve success rates of 92.31\%, 76.92\%, and 61.54\% for short, medium, and long routes, respectively, outperforming the best single-call baseline, Claude~3.7 Sonnet, with 84.62\%, 69.23\%, and 53.85\%, respectively. On MP-3, our workflow achieves success rates of 76.92\%, 61.54\%, and 38.46\%, improving over the best baseline of 61.54\%, 46.15\%, and 23.08\%. These results demonstrate the superiority of our method over single-call LLM baselines. Results on both the real-world floorplan and the CVC-FP benchmark further demonstrate that our workflow is a scalable solution for generating more accessible navigation for BLV individuals.

\end{abstract}    
\section{Introduction}
\label{sec:intro}
\begin{figure}[!t]
    \centering
    \includegraphics[width=0.7\columnwidth]{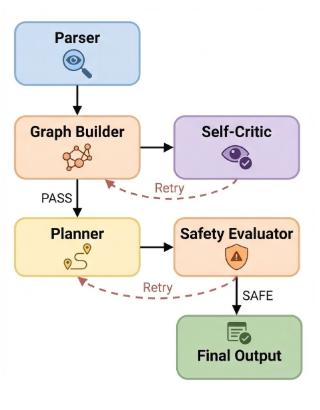}
  \caption{Overview of our agentic pipeline showing the five-agent workflow and three-tier RAG knowledge store.}
  \label{fig:schematic}
\end{figure}

Approximately 2.2 billion people live with some form of visual impairment~\cite{who2019}. While outdoor navigation is largely addressed by GPS-based smartphone applications, indoor wayfinding for blind and low-vision (BLV) users remains an open and consequential challenge. GPS signals attenuate inside buildings, and complex indoor topologies (e.g., hallways, doors, elevators) are difficult to communicate through audio alone~\cite{kuriakose2022indoor}. Existing indoor assistive systems generally follow one of two paradigms. Most approaches require deploying Bluetooth beacons or Wi-Fi fingerprinting, entail physical installation and maintenance in dynamic environments, and rely on vision-based methods that estimate depth from real-time camera feeds. These systems usually need to be retrained for each building. Although they can achieve accurate localization and support safe navigation, their dependence on environment-specific training limits their usability for users navigating among different indoor spaces. In our previous study \cite{ayanzadeh2025floorplan2guide}, we addressed this problem by transforming floor plans into a more navigable representation. Specifically, we create a structured, graphical representation of floor plan elements and convert the floor plan into a navigable system. This approach decreases preprocessing requirements while enabling the extraction of a traversable spatial graph directly from a single floor plan image and translating into navigation instructions. Our developed system takes a single floor plan image and a navigation query (start room and destination) as input and generates first-person, heading-tracked navigation steps enriched with landmark references, door identifiers, sensory confirmation cues, and safety warnings. Algorithm~\ref{alg:orchestrator} details the complete pipeline execution. The exact LLM prompts used by each agent are provided in the Appendix.

In this study, we extend our approach to improve the robustness of floor plan parsing by introducing an agentic pipeline that implements a hybrid Retrieval-Augmented Generation (RAG) framework. This system stores multiple perspectives of the floor plan, making the system more robust for navigation. Moreover, we employ fiducial markers~\cite{zafari2019survey,nakajima2023aruco} as navigation checkpoints. Since the primary focus of this research is floor plan parsing and the generation of safe navigation instructions, these markers perform as reliable reference points to support navigation without shifting the study's core emphasis.

\smallskip
\noindent Our contributions are summarized as follows:
\begin{enumerate}
    \item A Multi-agentic pipeline with self-correcting feedback loops for strong spatial graph extraction from architectural floor plan images.
    \item Integrate Retrieval-Augmented Generation (RAG) that uses multimodal information for grounded path planning which is safety-aware for BLV users.
    \item Evaluation on real-world public buildings and the CVC-FP benchmark, including physical walk-through validation, showing the effectiveness of the proposed method over single-call LLM baselines.

\end{enumerate}

The rest of this paper is structured as follows. Section~\ref{sec:related} reviews related work. Section~\ref{sec:method} details the proposed agentic pipeline and the hybrid RAG architecture. Section~\ref{sec:experiments} presents experiments and results on the CVC-FP benchmark and two real-world UMBC buildings~(MP-1 and MP-3), and Section~\ref{sec:conclusion} concludes the paper.


\begin{figure*}[!t]
    \centering
    \includegraphics[width=\textwidth]{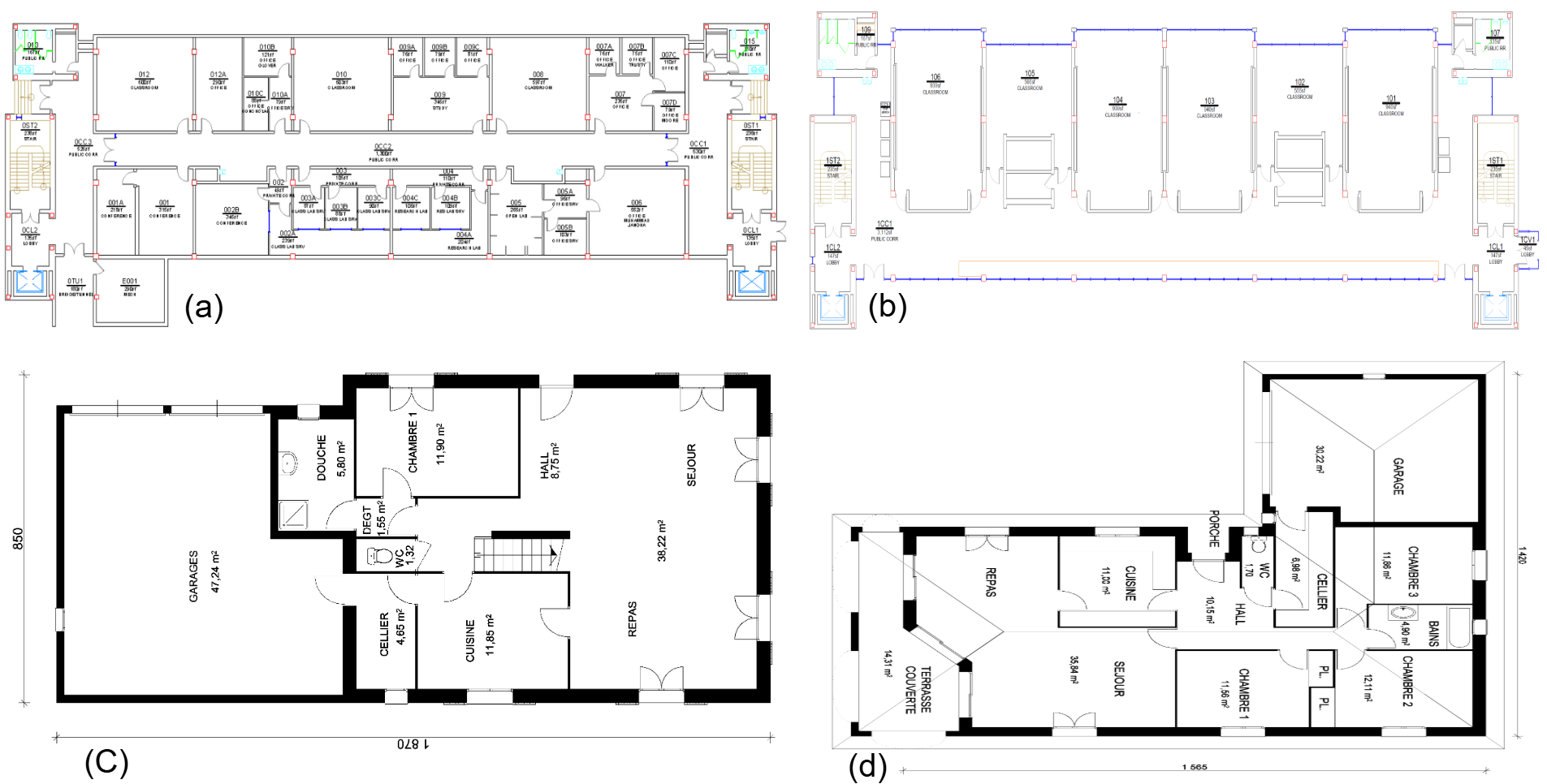}
    \caption{Sample floor plan images used in this study. (a) MP-3 and (b) MP-1 are architectural floor plans with detailed room and structural annotations. (c) and (d) are sample floor plans from the CVC-FP dataset.}
    \label{fig:floorplan_1}
\end{figure*}
\section{Related Work}\label{sec:related}

\subsection{Indoor Navigation for BLV Users}
Assistive indoor wayfinding has been investigated through both infrastructure-dependent and infrastructure-free paradigms.
Zafari~\emph{et al.}~\cite{zafari2019survey} survey Bluetooth Low Energy (BLE) beacon deployments, which achieve sub-meter accuracy but incur per-building installation costs.
Nakajima~\emph{et al.}~\cite{nakajima2023aruco} combine ArUco marker tracking with graph-based pathfinding for real-time guidance.
Vision-based approaches leverage depth estimation but degrade in textureless institutional corridors.
Kuriakose~\emph{et al.}~\cite{kuriakose2022indoor} comprehensively review smartphone-based solutions, concluding that no single modality provides the coverage and robustness needed for widespread BLV adoption. More recently, large language models (LLMs) have been harnessed for explicit spatial reasoning and navigation planning (e.g., NavGPT~\cite{zhou2024navgpt}, MapGPT~\cite{chen2024mapgpt}), occasionally leveraging multi-agent consultation to improve zero-shot routing~\cite{liang2024discussnav}.

\subsection{Floor Plan Parsing}

In our prior work, we explored building on graph embeddings and structural relationship capture for link prediction~\cite{ayanzadeh2020resvgae, nallbani2021resvgae} and improved floor plan parsing for BLV individuals by converting floor plans into a structured knowledge graph and leveraging foundation models for step-by-step navigation instructions~\cite{ayanzadeh2025floorplan2guide}. To our knowledge, this was among the first approaches to parse floor plans using an explicit graphical representation rather than relying solely on direct visual reasoning, as in prior Vision-Language and vision-based methods~\cite{li2024vlmparse}. Our results demonstrated the efficiency of structured floor plan parsing for Vision-Language Navigation (VLN) and greater robustness than direct visual prompting alone~\cite{ayanzadeh2025floorplan2guide}.

Coffrini~et~al.~\cite{coffrini2025llm} prompt Multimodal Language Models (MLLMs) to generate step-by-step indoor navigation instructions from floor plan images, showing that converting raw floor plans into graph-style schematics substantially improves spatial grounding and instruction accuracy. However, this structured representation is constructed manually, leaving automated extraction an open problem. ViLLA~\cite{singh2024villa} integrates LLMs with vision transformers for automatic floor plan analysis and feature extraction, while WAFFLE~\cite{yang2024waffle} uses LLMs and VLMs to structure textual metadata. Advances in spatial reasoning~\cite{chen2024spatialvlm} and large-scale architectural datasets~\cite{fan2021floorplancad} have similarly driven progress in floor plan vectorization and semantic generation~\cite{shabani2023housediffusion}. However, these approaches focus primarily on feature extraction or synthesis rather than evaluating layout accessibility for navigation. We build on these principles, introducing domain-specific agents, self-critics, and an evaluator to improve parsing accuracy and reduce hallucinations in path planning. Our approach is tailored to the spatial structure of floor plans and emphasizes safety, a critical factor for BLV users during indoor navigation.

\begin{figure*}[!t]
    \centering
    \includegraphics[width=\textwidth]{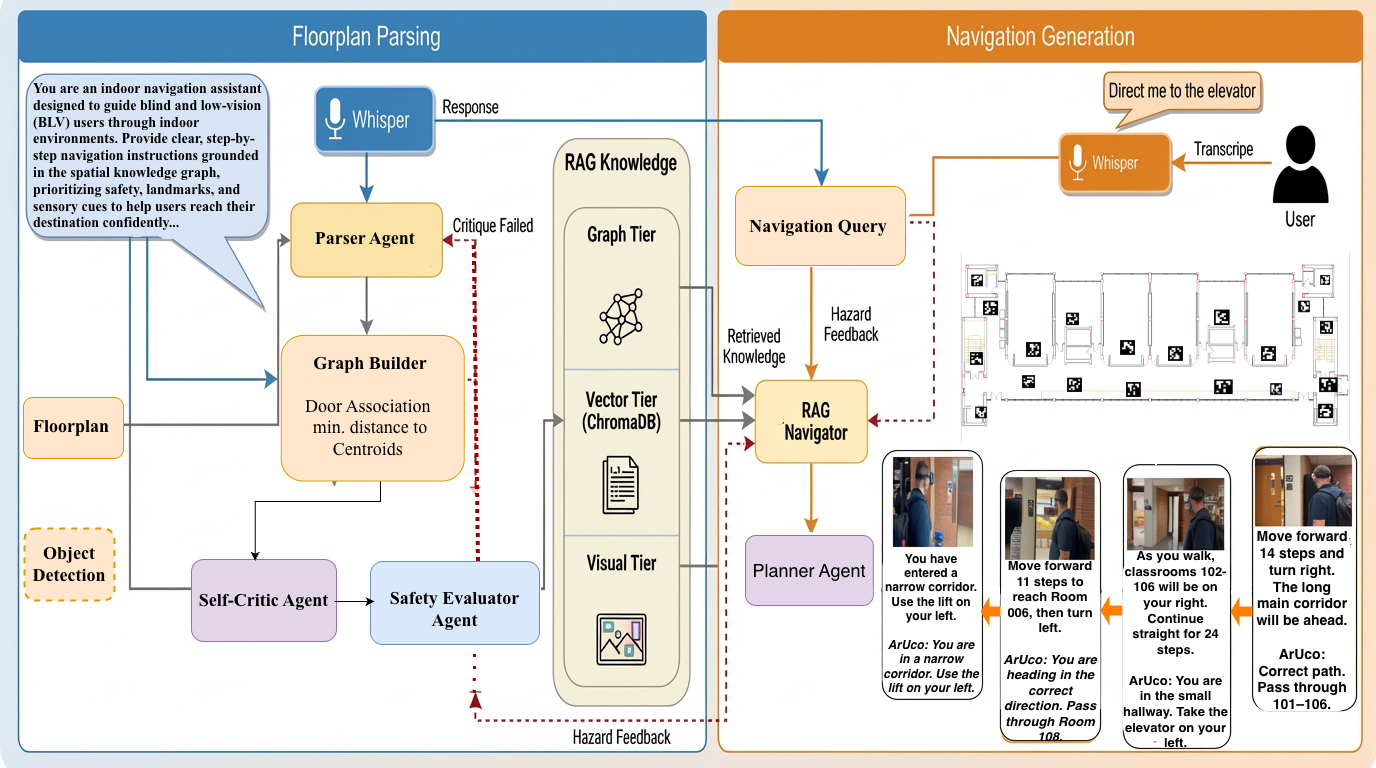}
    \caption{End-to-end architecture of the workflow. Left: The Parser extracts rooms and adjacencies, the Graph Builder constructs a validated floor-plan graph, and the Self-Critic enforces structural checks, triggering feedback as needed. A Safety Evaluator flags hazards such as narrow passages or dead ends. Center: Results are stored in a three-tier RAG structure (graph, vector, visual). Right: For a query, the RAG Navigator retrieves context, and the Planner generates BFS-based navigation with landmarks and safety notes, with re-routing on hazard alerts.}
    \label{fig:workflow}
\end{figure*}

\subsection{Retrieval-Augmented Generation}
Retrieval-Augmented Generation (RAG)~\cite{lewis2020rag} mitigates LLM hallucination by grounding generation in retrieved evidence. GraphRAG~\cite{edge2024graphrag} extends this paradigm to knowledge-graph-structured retrieval, demonstrating superior multi-hop reasoning.
UrbanKGent~\cite{ning2024urbankgent} applies LLM agents to urban knowledge graph construction.
While these works establish the power of structured retrieval, none have applied RAG to indoor spatial navigation from architectural documents.
Our three-tier architecture uniquely combines deterministic pathfinding with semantic and visual retrieval.

\subsection{Agentic Systems for Vision-Language Models}
Recent agentic systems adopt LLMs as embodied controllers for explicit reasoning and tool use~\cite{yao2023react, wang2023voyager}. Extending these paradigms to vision, frameworks such as MM-ReAct~\cite{yang2023mmreact} enable iterative multimodal reasoning over images and text. In this study, we build upon and extend the floor plan parsing paradigm by introducing an agentic RAG controller over the floor plan parsing pipeline. Specifically, we integrate a multi-step agent that retrieves relevant subgraphs and contextual information and performs hierarchical chain-of-thought reasoning, moving from global route planning to local turn-by-turn instructions. In contrast to prior VLM-based floor plan parsing systems that rely on few-shot prompting or fixed pipelines~\cite{li2024vlmparse,ganon2025waffle}, our agent can progressively refine, verify, and repair navigation plans. This enables more context-aware guidance that is consistent with the safety requirements for a solid workflow designed for BLV users~\cite{ayanzadeh2025floorplan2guide}.

\section{Method}\label{sec:method}

The system operates in two phases, as shown in Figure~\ref{fig:workflow}. First, we create a spatial knowledge graph for each building during the Knowledge Construction phase and store it in the RAG Knowledge base. During the Navigation phase, we retrieve a multi-tier context for a user’s query and produce step-by-step directions for navigating a building. Each floor plan image is treated as input to the complete agentic pipeline, implemented as a multi-agent structure: Parser $\rightarrow$ Graph Builder $\rightarrow$ Self-Critic $\rightarrow$ Planner $\rightarrow$ Safety Evaluator, while ingesting the corresponding RAG. To gain insight into the system's efficacy, we conduct a trial in which a sighted evaluator physically walked each route in the corresponding UMBC building(s) while listening to the generated instructions.

For each floor plan,  a knowledge graph is constructed in which each floor plan is represented as $\G = (\V, \E, \A)$, where $\V$ is the set of room nodes, $E \subseteq \V \times \V$ is the set of edges where each edge is mediated by a door or structural feature between two rooms in $\V$, and $A \in \{0,1\}^{N \times N}$ is a symmetric adjacency matrix, where $A[i][j] = 1$ if two rooms, $v_i$ and $v_j$, share a door or passage and $0$ otherwise. Each node $v$ in the graph contains the room name and semantic type (e.g., room, hallway, elevator, or stairs) as well as the centroid coordinates of the room obtained via OCR. We also store the estimated dimensions of each room along with an estimate of the OCR confidence. Each edge $E$ contains the structural element that mediates the edge (e.g., door) along with a bounding box around that element and an estimate for the traversal cost of traversing through the edge.

\subsection{Agentic Pipeline}\label{sec:agentic}

The agentic pipeline decomposes floor plan understanding into sequential agents, that managed by a central controller with retry logic, as shown in Figure~\ref{fig:workflow} (left) and Algorithm~\ref{alg:orchestrator}. The specific prompts used by each agent are detailed in the Appendix.

\paragraph{Agent~1: Parser.}
For inputting a floor plan image $I$ and a navigation query $(s, d)$, the Parser generates an initial graph estimate $\hat{\G}_0$ by integrating the image with object detection $\mathcal{D}$ with YOLOv12 as an object detection with class names, confidences and bounding boxes, and outputs the detected rooms, a symmetric relation matrix, and the information of the detected rooms. By incorporating detection positions into grounding context, the Parser avoids generating more adjacencies that are generated by LLMs.

\paragraph{Agent~2: Graph Builder.}
The Graph Builder processes the Parser output into a validated node-edge graph. For each detected door, the two nearest rooms, determined by pixel centroid distance, are linked. The entries which are not detected are preserved as a passage edge, and an adjacency matrix is rebuilt from the edge list to make sure both sides agree.

\paragraph{Agent 3: Self-Critic}
The Self-Critic enforces five structural checks: Breadth-First Search (BFS) connectivity, door-edge consistency, and spatial coherence. It flags centroid distances beyond $\mu + 2\sigma$. It also checks matrix symmetry and isolated nodes. Failures trigger the Orchestrator to re-invoke the Parser with the appended error context, up to $R_c = 2$ retries, as shown in Figure~\ref{fig:workflow}.
This mechanism, inspired by Self-Refine~\cite{madaan2023selfrefine}, enhances graph extraction accuracy and produces reliable navigation instructions.

\paragraph{Agent~4: Planner.}
The Planner computes the shortest path $P^* = \textrm{BFS}(\A, s, d)$ and generates heading-tracked, turn-by-turn instructions that incorporate landmark references from the detection data. Any existing safety annotations are included in the planning prompt.

\paragraph{Agent~5: Safety Evaluator.}
The Safety Evaluator applies four rule-based checks, including narrow passages (clearances less than 90 cm between walls and objects), missing door-edge flags, long traversals (traversing more than 5 steps or routing on a place with fewer landmarks, dead ends, and checking instructions that can the user to hit the wall, which can stem from hallucinating the LLMs. These hazards are potential safety concerns for visual grounding models that directly process a floor plan image together with natural language, with no preprocessing steps.  Each hazard is assigned a severity score $\sigma \in [1,5]$ along with a mitigation recommendation. If a hazard receives $\sigma \geq 4$, the system initiates a re-routing feedback loop to the Planner, as shown in Figure~\ref{fig:workflow}.

\begin{algorithm}
\caption{Agentic Pipeline for Floor Plan Parsing}
\begin{algorithmic}[1]
\Require Floor plan image $I$, query $(s, d)$, max retries $R_c$
\Ensure Navigation steps $\mathcal{N}$

\State $\mathcal{D} \leftarrow \textrm{YOLOv12}(I)$ \Comment{Object detection}

\For{$r = 0$ \textbf{to} $R_c$}
    \State $\hat{\G}_0 \leftarrow \textrm{Parser}(I,\,\mathcal{D})$
    \State $\G \leftarrow \textrm{GraphBuilder}(\hat{\G}_0,\,\mathcal{D})$
    \State $\mathit{passed} \leftarrow \textrm{SelfCritic}(\G,\,\mathcal{D})$
    \If{$\mathit{passed}$}
        \State \textbf{break}
    \EndIf
\EndFor

\State Ingest $\G$ and $\mathcal{D}$ into Knowledge Base

\State $P^* \leftarrow \textrm{BFS}(\A,\,s,\,d)$
\State $\mathcal{N} \leftarrow \textrm{Planner}(\G,\,P^*,\,s,\,d)$

\State $\sigma_{\max} \leftarrow \textrm{SafetyEval}(\G,\,\mathcal{N})$

\If{$\sigma_{\max} \geq 4$}
    \State $\mathcal{N} \leftarrow \textrm{Planner}(\G,\,P^*,\,s,\,d,\,\sigma_{\max})$
    \Comment{Re-route}
\EndIf

\State \Return $\mathcal{N}$
\end{algorithmic}
\label{alg:orchestrator}
\end{algorithm}

\subsection{Preprocessing}\label{sec:preproc}

Localizing room labels on the floor plans is difficult due to variable font styles, orientations, and background clutter. To address this, we adopt a multi-stage detection strategy that falls back via several methods until labels are successfully located. Specifically, we apply multi-engine OCR (e.g., PP-OCR~\cite{du2020ppocr}, EasyOCR, Tesseract) with fuzzy matching (Levenshtein ratio $\geq 0.55$) against known labels. We employ YOLOv12~\cite{tian2025yolov12}, fine-tuned to detect floor plan elements, including doors, walls, and windows. Each detection produces a node containing the class label, confidence score, bounding box, and center coordinates. These detections serve dual roles: grounding the parser’s LLM prompt and enabling the Graph Builder’s spatial door–room association.

\subsection{RAG Knowledge Base}\label{sec:rag}

After knowledge construction, the validated graph $\G$ and associated metadata are ingested into a persistent, queryable multimodal knowledge base, as depicted in the central panel of Figure~\ref{fig:workflow}.

\textbf{Graph-based Relational Knowledge:} The adjacency matrix and room metadata are loaded into a NetworkX~\cite{hagberg2008networkx} graph. Edge attributes include the mediating element, bounding box, and spatial descriptions for each node. Path queries use BFS, guaranteeing shortest paths in $O(|\V| + |\E|)$, providing the structural backbone for physically valid routes.

\textbf{Vector-based Semantic Knowledge:} For semantic embeddings, a rich text document is constructed for each room, door, and room-to-room transition, incorporating room name and type, dimensions, OCR confidence, neighboring rooms, accessible doors with VLM-generated context (wall side, spatial relationships), landmark descriptions, and surface materials. Documents are embedded and indexed in ChromaDB~\cite{chroma2024} for semantic retrieval:
$\textrm{doc}_i^* = \arg\max_j \textrm{sim}(\mathbf{q}, \mathbf{e}_j)$.
This modality transforms generic instructions (e.g., ``go to hallway'') into precise guidance (e.g., ``exit through Door~D2 on the south wall; the hallway has a smooth tile floor'').

\textbf{Visual Grounding Context:} The original image path, all YOLOv12 detections, and OCR label positions are stored. A VLM spatial analysis module generates context for each element (connected rooms, wall locations, adjacency), fusing visual grounding with the graph and vector modalities. Upon a query, the navigator retrieves: (1)~the shortest path from the graph store, (2)~rich semantic documents, and (3)~detection summaries and visual context. These are assembled into a structured prompt that guides the LLM along the exact graph path with landmark and sensory details. A post-generation cross-validation verifies that all referenced rooms exist on the computed path $P^*$; hallucinations or impossible transitions trigger instruction regeneration.

\subsection{Localization}
During navigation, fiducial markers serve as checkpoints. When a specific ArUco marker referenced in the navigation instructions is scanned, the system confirms the instruction. Otherwise, it sends an alert and reroutes the user from the detected checkpoint to the destination. Figure~\ref{fig:three-tier} shows a concrete example of the hybrid knowledge base populated from a CVC-FP floor plan dataset~\cite{CVC_FP}, illustrating the complementary information contributed by the graph, vector, and visual tiers. We conduct 101 navigation trials per route type, including short, medium, and long, to assess end-to-end system performance.



\begin{figure}[t]
  \centering
  \begin{tcolorbox}[colback=blue!3!white, colframe=blue!60!black,
    title=\textbf{Multimodal Knowledge Base (KB)},
    fonttitle=\sffamily\small, boxrule=0.5pt, left=4pt, right=4pt, top=4pt, bottom=4pt]
    
  {\scriptsize \textbf{Graph-based Relational Knowledge}}
  {\scriptsize
  \begin{verbatim}
Transition: Cuisine -> Repas
Via: door | Door ID: Door_D3
Door position (bbox): [160, 260, 180, 280]
Door description: Door_D3 to Repas
Cuisine size: 11.85 m2 (3.5 m x 3.4 m)
Repas size: 38.22 m2 (5.5 m x 6.95 m)
...
  \end{verbatim}
  }
  
  \tcbline
  
  {\scriptsize \textbf{Vector-based Semantic Knowledge}}
  {\scriptsize
  \begin{verbatim}
Room: Cuisine
Type: room | Size: 11.85 m2
Dimensions: 3.5 m x 3.4 m
Doors (3): Door_D2 to Cellier; Door_D3 to Repas;
           Door_D6 to Hall
Windows (1): West wall
Connected rooms: Cellier to North; Repas to South;
                 Hall to East
                 ....
  \end{verbatim}
  }
  
  \tcbline
  
  {\scriptsize \textbf{Visual Grounding Context}}
  {\scriptsize
  \begin{verbatim}
Image Source: Input Floor Plan Image
Detection: door | Confidence: 0.82
Bounding box: [685, 1993, 773, 2084]
Center: (729, 2038)
Room A: HALL | Room B: SEJOUR
Wall side: South wall of HALL
Door type: hinged single door
Description: Standard interior door connecting
  HALL to SEJOUR, on the south wall of HALL
 ...
  \end{verbatim}
  }
  \end{tcolorbox}
  \caption{Components of the multimodal knowledge base: graph-based relational knowledge, vector-based semantic knowledge, and visual grounding context.}
  \label{fig:three-tier}
\end{figure}

\section{Experiments}\label{sec:experiments}

\subsection{Setup}

We ran the system on a MacBook Pro equipped with an Apple M1 Pro chip, featuring a 10-core CPU, 16-core GPU, and 16 GB of unified memory. Live video was streamed from a head-mounted camera directly to the local system, where the ArUco detection module processed each frame to identify fiducial markers~\cite{garrido2014automatic} distributed throughout the environment. Each marker encodes a unique node ID, anchoring the user's position within the knowledge graph and enabling accurate routing to destinations. User queries are transcribed using Whisper~\cite{whisper}, simplifying the query process. Navigation performance is quantified using Success Rate~(SR), a standard benchmark metric defined as:
\begin{equation}
    SR = \frac{N_{\text{success}}}{N_{\text{total}}}
\end{equation}
where $N_{\text{success}}$ is the number of completed navigations and $N_{\text{total}}$ 
is the total number of trials.

\begin{figure}[!t]
    \centering
        \includegraphics[width=\columnwidth]{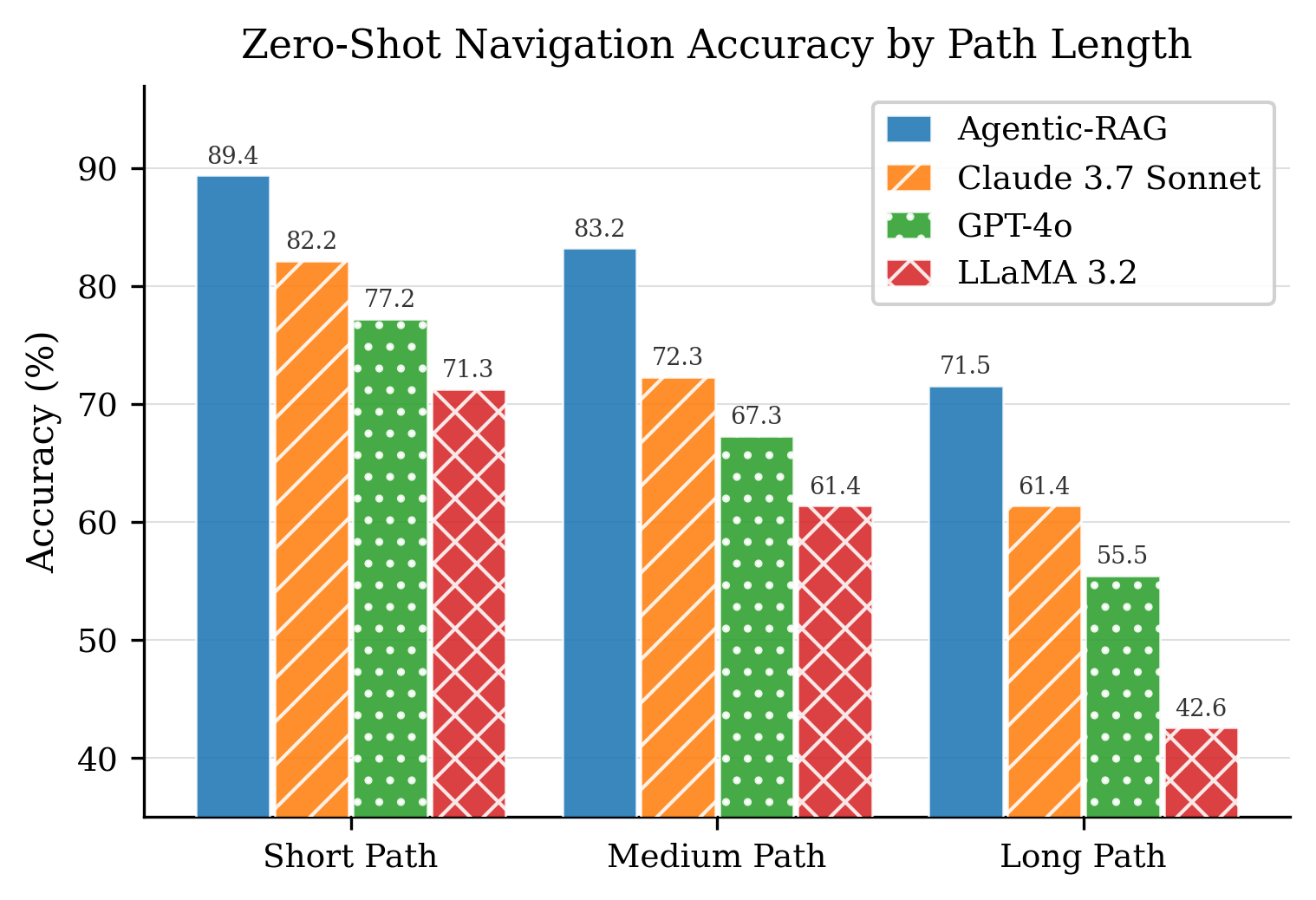}

  \caption{Bar plot comparing zero-shot navigation accuracy across short, medium, and long routes on CVC-FP for different models. }
  \label{fig:CVC-FP}
\end{figure}

\subsection{Evaluation on CVC-FP Benchmark}

We evaluate our proposed method using Claude~3.7~\cite{anthropic2024claude} Sonnet, GPT-4o, and LLaMA~3.2 Vision-Instruct~11B~\cite{llama3.2}, which we obtained from our prior study. A trial was considered successful when the system produced navigation instructions that accurately directed the user from the starting point to the destination without passing through structurally invalid transitions or entering inaccessible areas. As presented in Figure~\ref{fig:CVC-FP}, our system achieves the highest 
Success Rate~(SR) across all route types: 89.4\%, 83.2\%, and 71.5\% for short, medium, and long paths, respectively. These results surpass our prior baselines: 84.62\%, 69.23\%, and 53.85\% for Claude~3.7 Sonnet on short, medium, and long routes, respectively. LLaMA~3.2 records the lowest performance at 71.3\%, 61.4\%, and 42.6\% for short, medium, and long routes, respectively, on the CVC-FP dataset.

\subsection{Evaluation on MP Building}\label{sec:realworld}

To validate the system on real-world floor plans, a sighted evaluator physically walked 39 routes (13 short, 13 medium, 13 long) in the UMBC Math~\&~Psychology~(MP) building, following audio navigation instructions via screen reader. A route was marked successful if the evaluator reached the correct destination without deviating from the instructed path or encountering a safety hazard. Table~\ref{tab:MP1} reports results on MP-1, where the graph-augmented baselines (from our prior work~\cite{ayanzadeh2025floorplan2guide}) consistently outperform their graph-free counterparts. Our method achieves the best SR of 92.31\%, 76.92\%, and 61.54\% for short, medium, and long routes on MP-1, respectively. We further extend this evaluation to the larger and more complex MP-3 floor plan; results are reported in Table~\ref{tab:MP3_graph_ablation}. As shown in Table~\ref{tab:MP3_graph_ablation}, our system outperforms all baseline models on MP-3, achieving SR values of 76.92\%, 61.54\%, and 38.46\% for short, medium, and long routes, respectively. Claude~3.7 Sonnet reaches 61.54\% on short routes but falls to 23.08\% on long paths. GPT-4o achieves 61.54\% on short routes but degrades to 7.69\% on long routes. LLaMA~3.2 records the lowest performance, scoring 46.15\%, 15.38\%, and 7.69\% for short, medium, and long paths. Compared with MP-1, all models show decreased SR on MP-3 due to the larger number of rooms and increased layout complexity. Occasional OCR mislocalization degrades graph extraction quality and consequently navigation performance.

\begin{table}[htbp]
\centering
\caption{Success rate comparison on MP-1 in Zero-shot prompting,
with and without spatial knowledge graph augmentation.
Numbers in parentheses indicate successful trials out of 13 total
routes. Bold rows denote graph-augmented conditions.
Our method has no graph-free variant as the graph is a core
architectural component.}
\label{tab:MP1}
\footnotesize
\setlength{\tabcolsep}{4pt}
\renewcommand{\arraystretch}{1.10}
\begin{tabular}{|l|c|c|c|c|}
\hline
\textbf{Model} & \textbf{Graph} & \textbf{Short (\%)} & \textbf{Medium (\%)} & \textbf{Long (\%)} \\
\hline
\multirow{2}{*}{Sonnet 3.7~\cite{ayanzadeh2025floorplan2guide}}
  & $\times$ & 76.92 (10) & 61.54 (8)  & 38.46 (5) \\
  & \checkmark & \textbf{84.62 (11)} & \textbf{69.23 (9)}  & \textbf{53.85 (7)} \\
\hline
\multirow{2}{*}{GPT-4o~\cite{ayanzadeh2025floorplan2guide}}
  & $\times$ & 61.54 (8)  & 46.15 (6)  & 30.77 (4) \\
  & \checkmark & \textbf{69.23 (9)}  & \textbf{61.54 (8)}  & \textbf{46.15 (6)} \\
\hline
\multirow{2}{*}{LLaMA 3.2~\cite{ayanzadeh2025floorplan2guide}}
  & $\times$ & 46.15 (6)  & 23.08 (3)  & 15.38 (2) \\
  & \checkmark & \textbf{53.85 (7)}  & \textbf{38.46 (5)}  & \textbf{30.77 (4)} \\
\hline
\textbf{Our Method} & \checkmark & \textbf{92.31 (12)} & \textbf{76.92 (10)} & \textbf{61.54 (8)} \\
\hline
\end{tabular}
\end{table}



\begin{table}[htbp]
\centering
\caption{Success rate comparison on MP-3 in Zero-shot prompting,
with and without spatial knowledge graph augmentation.
Numbers in parentheses indicate successful trials out of 13 total
routes. Our approach has no graph-free variant as the graph is a
core architectural component.}
\label{tab:MP3_graph_ablation}
\footnotesize
\setlength{\tabcolsep}{4pt}
\renewcommand{\arraystretch}{1.10}
\begin{tabular}{|l|c|c|c|c|}
\hline
\textbf{Model} & \textbf{Graph} & \textbf{Short (\%)} & \textbf{Medium (\%)} & \textbf{Long (\%)} \\
\hline
\multirow{2}{*}{Sonnet 3.7}
  & $\times$ & 46.15 (6)  & 46.15 (6)  & 15.38 (2) \\
  & \checkmark & \textbf{61.54 (8)}  & \textbf{46.15 (6)}  & \textbf{23.08 (3)} \\
\hline
\multirow{2}{*}{GPT-4o}
  & $\times$ & 38.46 (5)  & 23.08 (3)  & 7.69 (1)  \\
  & \checkmark & \textbf{61.54 (8)}  & \textbf{38.46 (5)}  & \textbf{7.69 (1)}  \\
\hline
\multirow{2}{*}{LLaMA 3.2}
  & $\times$ & 23.08 (3)  & 7.69 (1)   & 0.00 (0)  \\
  & \checkmark & \textbf{46.15 (6)}  & \textbf{15.38 (2)}  & \textbf{7.69 (1)}  \\
\hline
\textbf{Our Method} & \checkmark & \textbf{76.92 (10)} & \textbf{61.54 (8)} & \textbf{38.46 (5)} \\
\hline
\end{tabular}
\end{table}

\subsection{Limitations and Future Work}In this study, we evaluate the proposed model on a single architectural layout; to extend this approach to multi-floor buildings will require integrating logic for elevator and staircase connectivity. Real-time localization, which combines user position data with floor plan navigation, remains unaddressed. This system relies on a fiducial marker~\cite{garrido2014automatic} as a localization checkpoint to assist users during navigation and does not use devices which are equipped with sensors such as LiDAR or IMU, which could enhance positioning accuracy and reduce rerouting associated with the exclusive use of ArUco markers. A preliminary real-world evaluation of UMBC buildings (Section~\ref{sec:realworld}) demonstrates that the system generates physically valid and navigable routes; however, the pilot study included only a single sighted evaluator. To address these limitations, a user study with BLV participants is required to evaluate the system's accessibility and usability in dynamic indoor environments. Moreover, in this study, we focus on the SR as the main metric by considering the importance of having accessible and safe navigation to reach the destination; to extend our study, besides the user study, we will include the different metrics, such as hallucination rate, rerouting metrics, and other related metrics, to measure the reliability of the proposed system. All future human-subject studies will be conducted under approved Institutional Review Board (IRB) protocols to ensure informed consent and participant safety.

\section{Conclusion}\label{sec:conclusion}

We present a system that converts a single architectural floor plan image into accessible, safety-aware indoor navigation instructions. Our agentic pipeline features self-correcting mechanisms to ease the robust generation of spatial knowledge graphs from the floor plan. The multimodal knowledge base grounds navigation generation in graph-based, semantic, and visual contexts and is employed by a planner agent, which is evaluated by a safety evaluator that has feedback loops acting as a built-in. We achieve navigation success rates of 92.31\% and 76.92\% on short routes in MP-1 and MP-3, respectively, improving over the best single-call baselines of 84.62\% (MP-1) and 61.54\% (MP-3). On the CVC-FP benchmark, we obtained 89.4\%, 83.2\%, and 71.5\% for short, medium, and long routes, consistently outperforming all single-call LLM baselines. This study demonstrates that large language models, when integrated within an agentic framework, can serve as a powerful backbone for floor plan parsing and accessible path planning, providing a scalable solution toward infrastructure-free, safe, and scalable indoor navigation for the BLV community.


\paragraph{Ethical Considerations.}

LLMs were used only to improve language clarity and readability. All research design, experimentation, and analytical decisions were performed independently by the authors.

\appendix
\clearpage

\twocolumn[{%
{\large\bfseries\sffamily Appendix~A\;}\\[2pt]
{
\\ Compact version of prompt for Parser, Planner, self-critics, and safety evaluator Agents.}
\vspace{12pt}
}]

\tcbset{
  promptstyle/.style={
    enhanced,
    colback=blue!2!white,
    colframe=cvprblue!85!black,
    colbacktitle=cvprblue!88!black,
    coltitle=white,
    fonttitle=\sffamily\bfseries\small,
    attach boxed title to top left={yshift=-2mm, xshift=5mm},
    boxed title style={
      colback=cvprblue!88!black,
      colframe=cvprblue!88!black,
      arc=2pt,
      boxrule=0pt,
      left=4pt, right=4pt, top=1.5pt, bottom=1.5pt,
    },
    boxrule=0.6pt,
    arc=3pt,
    left=5pt, right=5pt, top=8pt, bottom=4pt,
    breakable,
    before skip=6pt,
    after skip=6pt,
  }
}

\begin{tcolorbox}[promptstyle,
  title={\enskip Parser Agent}]
{\fontfamily{pcr}\selectfont\footnotesize\linespread{1.0}\selectfont
\begin{verbatim}
You are an AI analysing architectural
floor plan images. TASK: Extract
room-connectivity graph + metadata.
Return ONLY valid JSON.
{detection_context}

ADJACENCY RULE (CRITICAL): Two rooms
are adjacent ONLY IF they share a
direct door/passage. Transitive
connections via room C are forbidden:
use A--C and C--B instead of A--B.
USE detected door positions to
validate adjacency.
DUPLICATE NAMES: Number distinct areas
sharing a label (e.g. "Bathroom 1",
"Bathroom 2").
EVERY NODE MUST HAVE >= 1 EDGE.

JSON schema:
{ "approach": "<strategy>",
  "nodes_elements": [{"name":"<room>"}],
  "adjacency_matrix": [[0,1,...], ...],
  "edges": [{"from":"..","to":"..",
             "via":"<Door_D1|passage>",
             "door_bbox":[x1,y1,x2,y2]}],
  "rooms_info": [{"name":"..",
                  "size":"..","doors":[..],
                  "connected_rooms":[..]}]}
VALIDATION:
(1) len(nodes)==len(matrix)==len(row)
(2) matrix[i][j]==matrix[j][i] (symmetric)
(3) values in {0,1}, diagonal=0
(4) every node >= 1 edge
\end{verbatim}
}
\end{tcolorbox}

\begin{tcolorbox}[promptstyle,
  colframe=red!60!black,
  colbacktitle=red!65!black,
  boxed title style={colback=red!65!black,colframe=red!65!black,arc=2pt,
                     boxrule=0pt,left=4pt,right=4pt,top=1.5pt,bottom=1.5pt},
  title={\enskip Self-Critic Agent}]
{\fontfamily{pcr}\selectfont\footnotesize\linespread{1.0}\selectfont
\begin{verbatim}
You are a spatial reasoning expert
reviewing a floor plan graph for
structural errors.

GRAPH:  {graph_json}
EDGES:  {edges_json}
{detection_summary}
ISSUES ALREADY FOUND: {structural_issues}

CHECKLIST:
(1) All rooms that should be connected?
(2) Do edges match the detection data?
(3) Any rooms wrongly connected?
(4) Adjacency matrix consistent with
    edges?

Return ONLY JSON:
{ "passed": true/false,
  "issues": ["..."],
  "suggested_fixes": ["..."] }
\end{verbatim}
}
\end{tcolorbox}

\begin{tcolorbox}[promptstyle,
  colframe=green!45!black,
  colbacktitle=green!50!black,
  boxed title style={colback=green!50!black,colframe=green!50!black,arc=2pt,
                     boxrule=0pt,left=4pt,right=4pt,top=1.5pt,bottom=1.5pt},
  title={\enskip Planner Agent}]
{\fontfamily{pcr}\selectfont\footnotesize\linespread{1.0}\selectfont
\begin{verbatim}
Generate step-by-step navigation for a
visually impaired user.

GRAPH (validated): {graph_json}
EDGES:             {edges_json}
{safety_context}
TASK: {start} --> {destination}
      Step size: {step_size} cm.

HEADING (internal, not for user):
  N=0, E=90, S=180, W=270 deg.
  User starts facing start-room door.
  Turn left: heading-90;
  Turn right: heading+90.

LANDMARKS: Reference doors by ID
  (e.g."Door_D3"). Inject safety
  warnings if present.

OUTPUT JSON array:
[{ "step":               <int>,
   "action":             "<verb>",
   "heading_after_step": "<N|E|S|W>",
   "sensory_feedback":   "<feels/hears>",
   "current_position":   "<location>",
   "confirmation":       "<verify how>" }]
ALLOWED: Move forward N | Turn left |
         Turn right | Turn around | Stop
Final step MUST be Stop at destination.
\end{verbatim}
}
\end{tcolorbox}

\begin{tcolorbox}[promptstyle,
  colframe=orange!75!black,
  colbacktitle=orange!80!black,
  boxed title style={colback=orange!80!black,colframe=orange!80!black,arc=2pt,
                     boxrule=0pt,left=4pt,right=4pt,top=1.5pt,bottom=1.5pt},
  title={\enskip Safety Evaluator}]
{\fontfamily{pcr}\selectfont\footnotesize\linespread{1.0}\selectfont
\begin{verbatim}
You are a safety analyst reviewing a
route for a visually impaired user.

GRAPH:  {graph_json}
ROUTE:  {path}
EDGES:  {edges_json}
{detection_summary}
KNOWN HAZARDS: {existing_hazards}

ASSESS:
(1) Tight turns / complex manoeuvres?
(2) Risk of user disorientation?
(3) Sufficient landmarks for
    confirmation?
(4) Overall route complexity reasonable?

Return ONLY JSON:
{ "safe": true/false,
  "hazards": [{
    "hazard_type": "<type>",
    "severity":    <1--5>,
    "location":    "<where>",
    "mitigation":  "<how to mitigate>" }],
  "recommendation": "<overall>" }
\end{verbatim}
}
\end{tcolorbox}

\end{document}